\documentclass[letterpaper, 10 pt, conference]{ieeeconf}
\usepackage{textcomp}

\IEEEoverridecommandlockouts
\usepackage{gensymb}
\usepackage{textgreek}
\usepackage{graphicx}
\usepackage{amsmath}
\usepackage{cite}
\usepackage{booktabs}  
\usepackage{makecell}   
\usepackage[table]{xcolor}
\usepackage{colortbl}
\usepackage{url}
\usepackage{tabularx}
\usepackage{siunitx}
\usepackage{amssymb}
\begin{document}

\title{\LARGE \bf
Learning Hip Exoskeleton Control Policy via Predictive Neuromusculoskeletal Simulation}

\author{Ilseung Park, Changseob Song, and Inseung Kang,~\IEEEmembership{Member,~IEEE}
\thanks{This research was supported by the NIH R21 Award 1R21EB037268-01  (corresponding author: Ilseung Park, {\tt\footnotesize ilseungp@andrew.cmu.edu})}
\thanks{I. Park, C. Song, and I. Kang are with the Department of Mechanical Engineering, Carnegie Mellon University, Pittsburgh, PA, 15213 USA.}
}

\maketitle
\begin{abstract}
Developing exoskeleton controllers that generalize across diverse locomotor conditions typically requires extensive motion-capture data and biomechanical labeling, limiting scalability beyond instrumented laboratory settings. Here, we present a physics-based neuromusculoskeletal learning framework that trains a hip-exoskeleton control policy entirely in simulation, without motion-capture demonstrations, and deploys it on hardware via policy distillation. A reinforcement learning teacher policy is trained using a muscle-synergy action prior over a wide range of walking speeds and slopes through a two-stage curriculum, enabling direct comparison between assisted and no-exoskeleton conditions. In simulation, exoskeleton assistance reduces mean muscle activation by up to 3.4\% and mean positive joint power by up to 7.0\% on level ground and ramp ascent, with benefits increasing systematically with walking speed. On hardware, the assistance profiles learned in simulation are preserved across matched speed-slope conditions ($r$: 0.82$\pm$0.19, RMSE: 0.03$\pm$0.01~Nm/kg), providing quantitative evidence of sim-to-real transfer without additional hardware tuning. These results demonstrate that physics-based neuromusculoskeletal simulation can serve as a practical and scalable foundation for exoskeleton controller development, substantially reducing experimental burden during the design phase.

\end{abstract}

\vspace{0.2cm}


\section{Introduction}
Lower-limb wearable robots, such as exoskeletons, have demonstrated considerable potential to enhance human mobility by reducing metabolic cost or muscular effort during locomotion \cite{sawicki2020exoskeleton, siviy2023opportunities, Gao2025}. Beyond augmentation in able-bodied individuals, these systems hold significant promise for individuals with motor impairments where they can facilitate rehabilitation and improve functional walking capacity \cite{lerner2017lower, awad2017soft, kim2024soft, gunnell2025powered, kang2025online, pruyn2026portable}. Despite these advances, widespread real-world deployment remains limited. A central challenge is exoskeleton control: the assistance strategy that maximally benefits the user is not known \textit{a priori}, as the optimal solution is inherently dependent on the locomotor task.

Human-in-the-loop optimization has therefore been widely adopted to tune parameterized assistance torque profiles towards gait performance objectives such as minimizing metabolic cost \cite{slade2024human}. While this method has yielded substantial performance improvements \cite{zhang2017human, ding2018human, witte2020improving, kim2022reducing}, the resulting controllers are typically optimized for a fixed task and do not readily generalize to the diverse terrains, speeds, and task transitions encountered in daily life. To address this limitation, Molinaro \textit{et al}. recently introduced a task-agnostic control paradigm that modulates exoskeleton assistance based on a continuously estimated physiological state derived from onboard sensors \cite{molinaro2024task, molinaro2024estimating}. Trained on a broad dataset of human biomechanics, their controller directly maps estimated biological joint moment to exoskeleton torque commands, enabling generalization across a wide range of lower-limb activities without explicit task classification (both cyclic locomotion and non-cyclic transient movements).

\begin{figure*}[t!]
    \centering
    \includegraphics[width=\textwidth]{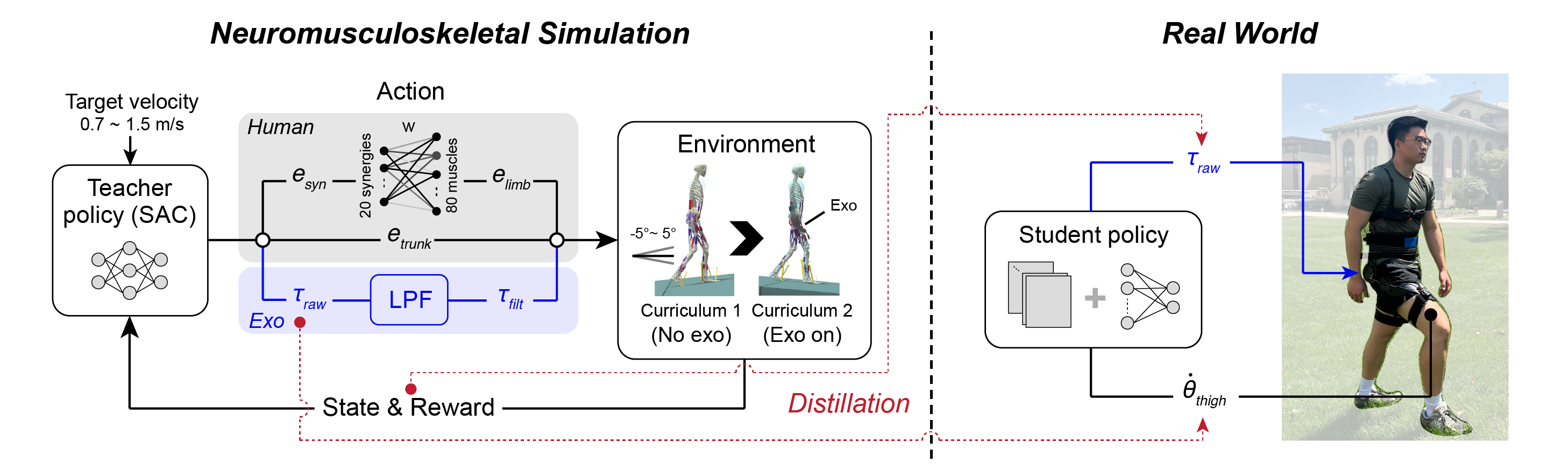}
\caption{\textbf{Simulation-to-real workflow for learning and deploying hip-exoskeleton control policy.} 
\textbf{Left (Neuromusculoskeletal simulation):} a privileged teacher policy is trained in a predictive human simulation across randomized target speeds ($0.7$--$1.5$~m/s) and slopes ($-5^\circ$ to $+5^\circ$) using a two-stage curriculum: \emph{Curriculum 1} learns stable locomotion without exoskeleton actuation (no-exoskeleton), then \emph{Curriculum 2} introduces bilateral hip exoskeleton actuation (exoskeleton-assisted), enabling the policy to learn assistive torques. The teacher outputs reduced-dimensional actions via a muscle-synergy prior, comprising lower-limb muscle excitations ($e_{\mathrm{syn}}\!\rightarrow\! e_{\mathrm{limb}}$), direct trunk-muscle excitations ($e_{\mathrm{trunk}}$), and raw bilateral hip-exoskeleton torques ($\tau_{\mathrm{raw}}$), which are smoothed by a first-order low-pass filter (LPF) to yield applied torques ($\tau_{\mathrm{filt}}$). Simulation state and reward are fed back to optimize the teacher.
\textbf{Right (Real world):} the teacher is distilled into a temporal convolutional network student policy that maps a short history of unilateral thigh inertial measurement unit (IMU) gyroscope signals ($\dot{\theta}_{\mathrm{thigh}}$) to unilateral hip-torque commands for onboard control; the resulting $\tau_{\mathrm{raw}}$ is passed through the onboard LPF and applied to the physical exoskeleton. The dashed red arrows denote the distillation link from simulated rollouts to the deployable IMU-only policy.}

    \label{framework}
\end{figure*}
\begin{figure}[t!]
    \centering
    \includegraphics[width=\columnwidth]{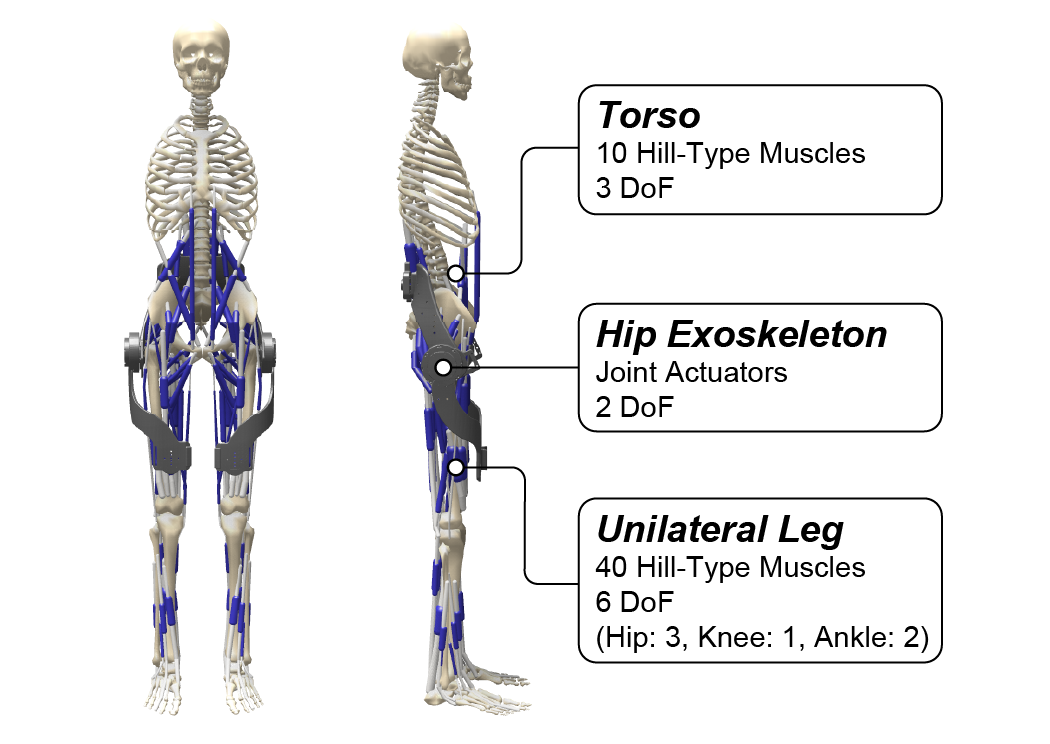}
    \caption{\textbf{Human-exoskeleton musculoskeletal model used in simulation.} Frontal and lateral view of the H2190 full-body musculoskeletal model, comprising 21 degrees of freedom and 90 Hill-type musculotendon actuators. Bilateral hip exoskeleton actuators apply flexion-extension torques about the left and right hip joints.}
    \label{model}
\end{figure}

While this task-agnostic paradigm represents the current state of the art, critical limitations remain. As with most data-driven, supervised learning-based controllers, performance is ultimately bounded by the coverage of the training distribution and can degrade under out-of-distribution conditions (e.g., unseen slopes and gait patterns) \cite{Kang2025, Scherpereel2025}. This challenge is further amplified in controllers that rely on estimating physiological states, such as joint moments, because these target variables are not directly observable. Obtaining ground-truth labels requires synchronized three-dimensional kinematics and ground reaction force (GRF) measurements. Consequently, control policy development remains largely confined to controlled lab environments, creating a critical bottleneck for scaling these approaches to diverse tasks and user populations.

To alleviate the reliance on manual data collection and instrumented experimentation, reinforcement learning in physics-based simulation offers a compelling and complementary path for exoskeleton control development \cite{badie2025bioinspired, leem2026exo}. Training entirely in simulation enables the efficient generation of large, diverse locomotor experience, allowing policies to encounter a broad range of gait patterns, terrains, and perturbations that would be difficult or infeasible to obtain through human experiments alone. Simulation further enables systematic evaluation under conditions that are impractical or unsafe to impose on human participants directly, including extreme slopes, large balance perturbations \cite{Ma2025CoRL}, and highly fatiguing workloads \cite{Chang2001}.

Although simulation-trained controllers have achieved strong sim-to-real transfer in several robotic domains, including legged locomotion \cite{ha2025learning}, comparatively few studies have developed and deployed closed-loop exoskeleton control policies derived from physics-based human simulation onto physical hardware. Prior work in assistive exosuits has predominantly leveraged musculoskeletal simulation to inform or hand-design candidate assistance moment profiles, which were subsequently validated experimentally rather than learned end-to-end and deployed as a closed-loop policy directly on a an onboard device \cite{kim2019reducing}.

To our knowledge, only one study to date has demonstrated full sim-to-real transfer by training an assistance policy in simulation and deploying it on a physical lower-limb exoskeleton, with validation of real-world locomotion performance \cite{luo2024experiment}. However, two critical limitations remain. First, the simulation-side evaluation was incomplete: the study did not establish whether the learned controller meaningfully reduced user effort within simulation, nor did it quantitatively benchmark simulated kinematics and kinetics against experimental human data to assess biomechanical fidelity. Second, the sim-to-real gap was not explicitly characterized: it remains unclear whether the assistance strategy learned in simulation was preserved on the physical device, as systematic comparisons of kinematics, kinetics, and resulting assistance profiles between the simulated and real-world conditions were not reported.
\begin{table}[t]
  \centering
  \caption{SAC hyperparameters and curriculum schedule.}
  \label{tab:sac_hyperparams}
  \setlength{\tabcolsep}{4pt}
  \renewcommand{\arraystretch}{1.05}
  \begin{tabularx}{\columnwidth}{@{}l X@{}}
    \toprule
    Parameter & Value \\
    \midrule
    Algorithm & Soft Actor--Critic (SAC) \\
    Implementation & Stable-Baselines3 \\
    Policy & Multilayer perceptron \\
    Actor/Critic sizes & [512, 512, 256] \\
    Parallel envs & 30 \\
    \midrule
    Stage 1 steps & $0$--$5\times10^{7}$ \\
    Exo (stage 1) & Zero-clamped \\
    LR (stage 1) & Linear decay $10^{-3}\!\rightarrow\!0$ \\
    \midrule
    Stage 2 steps & $5\times10^{7}$--$1\times10^{8}$ \\
    Exo (stage 2) & (i) Policy-controlled, (ii) zero-clamped \\
    LR (stage 2) & Linear decay $5\times10^{-4}\!\rightarrow\!0$ \\
    \midrule
    Discount $\gamma$ & 0.99 \\
    Replay buffer & $3\times10^{6}$ \\
    Batch size & 512 \\
    Target update $\tau_{\mathrm{target}}$ & 0.005 \\
    Entropy coeff. & auto\_0.1 \\
    \bottomrule
  \end{tabularx}
\end{table}
In this work, we address these limitations by proposing a sim-to-real learning pipeline for hip-exoskeleton control that requires no motion-capture demonstrations. We develop a predictive neuromusculoskeletal walking simulation augmented with bilateral hip actuators, and within this environment, train a reinforcement learning agent to coordinate muscle excitations and exoskeleton torques across a wide range of walking speeds and slopes using a muscle-synergy action prior \cite{bernstein1967coordination, bizzi2013neural}. Training follows a two-stage curriculum: the agent first learns stable unassisted locomotion, then we introduce exoskeleton actuation, enabling direct comparison against a matched no-exoskeleton baseline. The resulting teacher policy is subsequently distilled into an inertial measurement unit (IMU)-only student policy that maps a short history of thigh gyroscope signals to real-time hip-torque commands for onboard deployment.

We validate simulation fidelity by benchmarking simulated sagittal-plane joint angles and net joint moments against open-source human biomechanics data across multiple speeds and slopes. We then quantify the sim-to-real gap by comparing assistance torque waveforms between simulation and hardware over matched speed-slope conditions. Together, our approach provides an end-to-end pathway from physics-based simulation to embedded exoskeleton control, furnishing quantitative evidence of both biomechanical simulation validity and the preservation of the learned assistance strategy across sim-to-real transfer.

\section{Methods}
We developed a neuromusculoskeletal simulation framework using a two-stage curriculum (Fig. ~\ref{framework}). In the first stage, we trained a muscle-synergy-based controller to produce stable walking across a wide range of locomotor conditions, including speeds from $0.7$ to $1.5$~m/s and terrain inclinations from $-5^\circ$ to $+5^\circ$. In the second stage, we incorporated a robotic hip exoskeleton model, capable of delivering bilateral assistance torques up to $12$~Nm, and trained the exoskeleton assistance policy under the same locomotor curriculum. We also trained a matched no-exoskeleton condition for an identical number of environment steps to serve as a controlled baseline for evaluating the effect of exoskeleton assistance.

The resulting simulation-trained policy, referred to as the teacher, relies on privileged full simulator state information and is therefore not directly deployable on embedded hardware. To bridge this gap, we distilled the teacher into a deployable student controller via behavioral cloning \cite{osa2018algorithmic}. The student policy receives a short history of bilateral femur mediolateral angular velocity as input and outputs real-time hip torque commands, enabling onboard deployment and hardware-based experimental evaluation of the learned assistance strategy.

\subsection{Neuromusculoskeletal Simulation}
Simulations were conducted in the Hyfydy physics engine integrated with the SCONE framework~\cite{Geijtenbeek2019,Geijtenbeek2021Hyfydy}. The control policy operated at $40$~Hz, while the forward dynamics were integrated at $200$~Hz.

\subsubsection{Human-Exoskeleton Model}
We used the H2190 full-body musculoskeletal model with the upper limbs removed, yielding a system with 21 degrees of freedom and 90 muscles~\cite{schumacher2025emergence} (Fig.~\ref{model}). Of these, 80 muscles actuated the lower extremities and the remaining 10 spanned the pelvis and torso. Each muscle was modeled as a Hill-type musculotendon actuator governed by first-order activation dynamics.

To represent the hip exoskeleton in the second curriculum stage, we added bilateral torque actuators to apply flexion-extension torques about the left and right hip joints. We accounted for the physical mass of the device by augmenting the segment inertial properties of the model: $0.5$~kg was added to each thigh, $2.0$~kg to the pelvis, and $1.3$~kg to the torso, for a total mass of $4.3$~kg. Segment inertial properties were scaled proportionally to the added mass for each segment. The resulting model comprised 90 muscle actuators and two exoskeleton actuators.

To avoid unrealistically rapid changes in the delivered assistance torque, the raw actuator commands were smoothed through a discrete-time first-order causal low-pass filter. The filter update at each time step $k$ is given by

\begin{equation}
u_k = u_{k-1} + \alpha \bigl(u_{\mathrm{cmd},k} - u_{k-1}\bigr)
\label{eq:lpf_discrete}
\end{equation}

where $u_{\mathrm{cmd},k}$ is the raw torque command, $u_k$ is the filtered actuator output, and $\alpha$ is the smoothing coefficient defined as

\begin{equation}
\alpha = \frac{\Delta t}{\tau_{\mathrm{LPF}}}
\label{eq:alpha}
\end{equation}

Here, $\Delta t$ is the simulation time step and $\tau_{\mathrm{LPF}}$ is the filter time constant. In this study, $\tau_{\mathrm{LPF}}$ was set to $0.1$~s, and the peak actuator torque magnitude was constrained to $12$~Nm.

\subsubsection{Policy Training and Curriculum}
We trained stochastic control policies using the Soft Actor--Critic (SAC) algorithm implemented in Stable-Baselines3~\cite{stable-baselines3}. Training followed a two-stage curriculum. In the first stage ($0$ to $5\times10^{7}$ environment steps), the policy controlled only the musculoskeletal actuators; exoskeleton torque commands were clamped to zero, allowing the agent to establish stable unassisted locomotion across the full range of target speeds and slopes. In the second stage ($5\times10^{7}$ to $1\times10^{8}$ steps), training proceeded under two parallel conditions: (1) an exoskeleton-assisted condition, in which the policy additionally controlled bilateral hip torque actuators, and (2) a matched no-exoskeleton condition, in which exoskeleton commands remained clamped to zero for an equivalent number of training steps. To stabilize learning at the transition between stages, we reset the learning-rate schedule to decay linearly from $5\times10^{-4}$ to $0$ over the remaining steps; all other SAC hyperparameters were held constant across both stages and conditions (Table~\ref{tab:sac_hyperparams}).

\subsubsection{Muscle Synergy Action Prior}
Directly controlling individual muscle actuators poses a high-dimensional, weakly constrained optimization problem that can produce physiologically implausible activation patterns and impede policy learning. To impose a structured, low-dimensional action prior, we derived a muscle-synergy representation from human overground walking data. Whole-body kinematics and GRFs were recorded, filtered, and segmented into individual strides. Muscle activations were then estimated by tracking the experimental motion using an OpenSim-converted H2190 model with OpenSim Moco (MocoInverse)~\cite{dembia2020opensim}, which enforces consistency with the measured kinematics and kinetics while minimizing muscle effort. Non-negative matrix factorization (NMF) was subsequently applied to the right-leg activation matrix to obtain a low-rank decomposition ~\cite{daniel1999learning}, yielding a fixed synergy weight matrix that parameterizes lower-limb muscle activations throughout the controller. The resulting policy action space comprised 32 control inputs: 10 synergy coefficients per leg (20 total), 10 direct activation signals for the torso and pelvis muscles, and two bilateral hip exoskeleton torque commands. In the no-exoskeleton condition, the exoskeleton commands were clamped to zero; in the exoskeleton-assisted condition, the policy directly specified the left and right hip torque outputs.

\subsubsection{Action and Observation Space}
The teacher policy was trained with privileged access to a rich, simulation-based observation space. At each time step, the observation vector comprised muscle fiber lengths, velocities, forces, and excitations for all actuated muscles; joint positions and velocities; head orientation and angular velocity; the relative positions of both feet; and GRFs under each foot, all expressed in a normalized form. In addition, the observation included the current bilateral hip exoskeleton torques (normalized by the maximum torque magnitude), the whole-body center-of-mass velocity, and the target walking speed. Together, this privileged state representation enabled the teacher policy to jointly coordinate muscle activations and exoskeleton torques based on complete musculoskeletal and device state information.

\subsubsection{Domain Randomization}
To expose the policy to diverse locomotor conditions, we randomized both the ground slope and the target walking speed at the start of each training episode. Eleven discrete terrain slopes, ranging from $-5^\circ$ to $+5^\circ$ in $1^\circ$ increments, were sampled from a categorical distribution whose probabilities were governed by per-slope difficulty scores. These scores were initialized uniformly and updated after each episode: a fall on a given slope increased its score, while successful episode completion decreased the score, bounded below by a minimum value. Slope sampling was thus biased toward conditions on which the policy still performed poorly, while periodically revisiting easier terrain to maintain stable locomotion. In parallel, the target walking speed followed a cyclic curriculum: a predefined speed sequence increased from $0.7$~m/s to $1.5$~m/s and then decreased back to $0.7$~m/s, applied repeatedly across episodes throughout training \cite{chiu2025learning}.

\subsubsection{Reward design}
At each simulation step $t$, the agent received a scalar reward composed of weighted component terms:

\begin{equation}
\begin{aligned}
    r_t ={}& w_{\text{vel}}\,r^{\text{vel}}_t 
           + w_{\text{eff}}\,r^{\text{eff}}_t 
           + w_{\text{rom}}\,r^{\text{rom}}_t \\
           &+ w_{\text{sm}}\,r^{\text{sm}}_t
           + w_{\text{fall}}\,r^{\text{fall}}_t
           + w_{\text{knee}}\,r^{\text{knee}}_t
           + w_{\text{front}}\,r^{\text{front}}_t
\end{aligned}
\end{equation}

where each $w_{(\cdot)}$ denotes the scalar weight assigned to the corresponding reward component $r^{(\cdot)}_t$. The velocity term $r^{\text{vel}}_t$ encouraged walking at the target forward speed while maintaining stable whole-body posture, implemented as the product of Gaussian functions of forward and vertical center-of-mass (COM) velocity, head orientation, and head angular velocity, with a flat region around the target speed to permit a small tolerance. The effort term $r^{\text{eff}}_t$ penalized the sum of squared muscle activations, discouraging energetically expensive coordination strategies. The range-of-motion term $r^{\text{rom}}_t$ penalized excessive knee flexion beyond $0^\circ$ and lumbar extension angles outside the interval $[-30^\circ, 2.5^\circ]$, enforcing physiologically plausible joint kinematics. The knee-load term $r^{\text{knee}}_t$ penalized stance-phase knee joint loads exceeding $3$ bodyweights, promoting load distribution across the limbs. The frontal-plane term $r^{\text{front}}_t$ penalized squared hip adduction angles bilaterally, reducing excessive lateral motion and circumduction. The smoothness term $r^{\text{sm}}_t$ penalized the squared finite difference of successive left and right exoskeleton torque commands, discouraging abrupt changes in delivered assistance. Finally, $r^{\text{fall}}_t$ applied a one-time large negative penalty when the COM height dropped below a predefined relative threshold, penalizing fall events.

\subsection{Policy Distillation}
\begin{figure}[t!]
    \centering
    \includegraphics[width=\columnwidth]{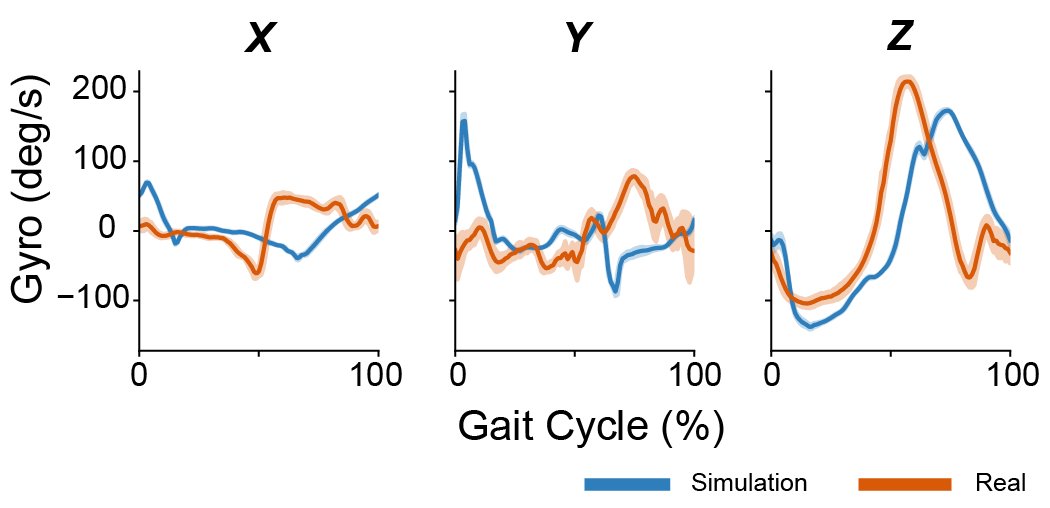}
    \caption{\textbf{Comparison of thigh gyroscope signals between simulation and hardware during level-ground walking at $1.2$~m/s.} Simulation signals were computed as femur angular velocity expressed in the femur local frame; hardware signals were recorded from an IMU mounted on the distal thigh bar of the exoskeleton. The rotational axes were defined with respect to an upright standing posture: $x$ corresponds to the anteroposterior axis, $y$ to the vertical axis, and $z$ to the mediolateral axis. Solid lines show stride-averaged means and shaded regions indicate $\pm 1$~SD across strides. The mediolateral component (gyro $z$) exhibited the strongest agreement ($r=0.55$) and was therefore selected as the sole input modality for student policy training.}

    \label{sensor_compare}
\end{figure}

\begin{figure*}[t!]
    \centering
    \includegraphics[width=\textwidth]{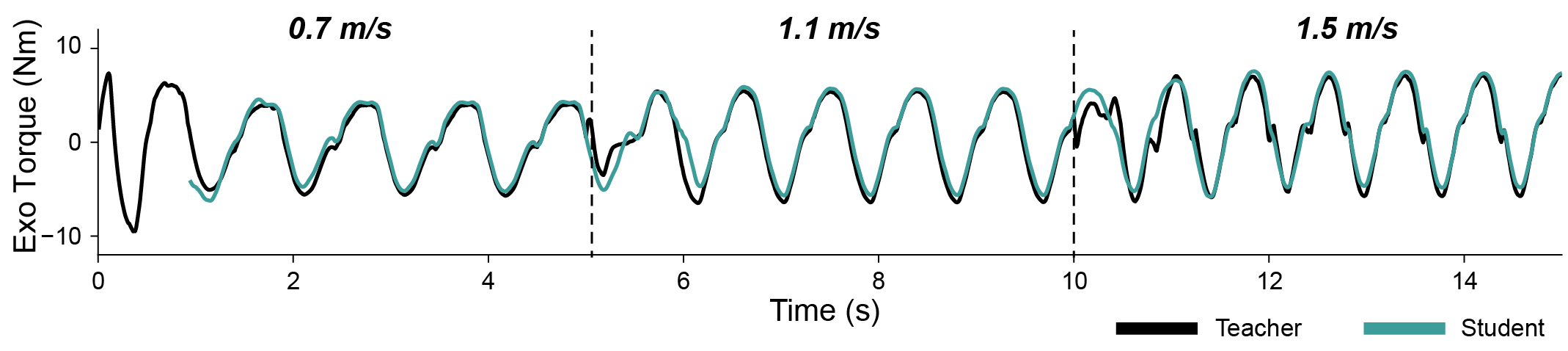}
\caption{\textbf{Teacher-student agreement in hip torque prediction.} Representative time series of right hip assistance torque produced by the privileged teacher policy (black) and the distilled student controller (teal). The student is a temporal convolutional network that predicts hip torque from a short history of femur mediolateral gyroscope measurements expressed in the local segment frame. The example trial was collected on level ground while commanding a sequence of target speeds ($0.7$, $1.1$, and $1.5$~m/s), each maintained for $5$~s. The first $0.95$~s of the trial contains no student output, as the input history window has not yet been fully populated. For this trial, the student closely tracked the teacher output ($R^2=0.93$).}

    \label{distillation}
\end{figure*}
\begin{figure}[t!]
    \centering
    \includegraphics[width=\columnwidth]{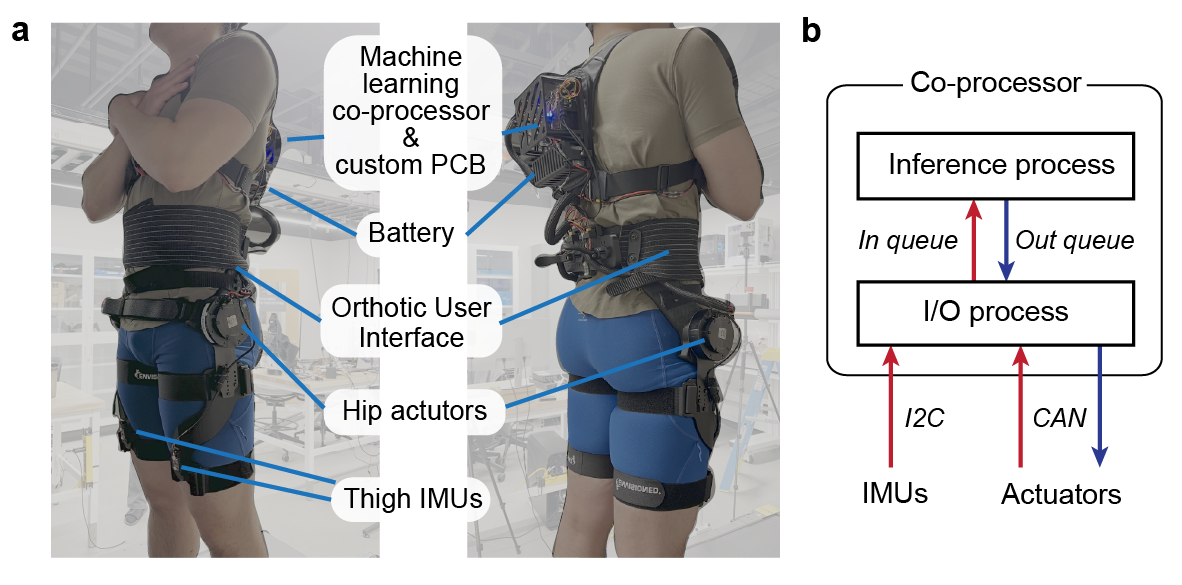}
        \caption{\textbf{Robotic hip exoskeleton hardware and onboard software architecture.} \textbf{(a)} Robotic hip exoskeleton designed to apply bilateral hip flexion-extension torques during locomotion. \textbf{(b)} Onboard software architecture of the machine learning co-processor. Input sensor signals are logged via a dedicated I/O process; a parallel inference process generates real-time hip torque commands using the pre-trained student policy.}

    \label{Exo_hw_sw}
\end{figure}

To deploy the learned assistance policy on hardware with limited sensing, we distilled the privileged teacher policy into a student policy that operates on a single wearable-sensor modality. To identify which IMU channel exhibited the strongest agreement between simulation and hardware, thigh gyroscope signals were compared during level-ground walking at $1.2$~m/s. In simulation, the femur angular velocity was computed in the femur local frame; on hardware, gyroscope data were recorded from an IMU mounted on the distal thigh bar of the exoskeleton while a participant walked for one minute at $1.2$~m/s. As shown in Fig.~\ref{sensor_compare}, the mediolateral component (gyro $z$) exhibited the strongest agreement between simulation and hardware (RMSE $=96.0$~deg/s, correlation $r=0.55$), whereas the remaining axes showed poor agreement (gyro $x$: RMSE $=41.2$~deg/s, $r=-0.19$; gyro $y$: RMSE $=60.9$~deg/s, $r=-0.25$). Based on this analysis, the mediolateral gyroscope signal (gyro $z$) was selected as the sole input modality for student policy training.

We generated training data by rolling out the teacher policy across three terrain conditions: level ground, ramp ascent ($+5^\circ$), and ramp descent ($-5^\circ$), over a range of target walking speeds. During rollouts, the controller operated at $100$~Hz. At each time step, the teacher's instantaneous right hip torque commands were paired with a $0.95$~s history window (95 samples) of the right femur mediolateral angular velocity, yielding supervised learning samples consistent with realistic onboard sensing. The dataset included both constant-speed trials and trials with smoothly varying target speeds.

The student policy was implemented as a temporal convolutional network (TCN) that maps a single-channel IMU signal history (mediolateral angular velocity) to the teacher’s right hip torque command. The network was trained for five epochs using a mean squared error loss, with a held-out validation set to monitor generalization. Training loss decreased from $7.45\times10^{-2}$ to $4.99\times10^{-2}$ on the training set and from $8.23\times10^{-2}$ to $6.54\times10^{-2}$ on the validation set (normalized torque units, $[-1,1]$). Fig.~\ref{distillation} shows a representative level-ground trial with step changes in target speed ($0.7$, $1.1$, and $1.5$~m/s; $4$~s each), where the student closely tracked the teacher output ($R^2=0.93$), demonstrating the feasibility of real-time deployment without access to the full simulator state.

\subsection{Robotic Hip Exoskeleton}

We developed a robotic hip exoskeleton to apply bilateral hip torques during overground locomotion (Fig.~\ref{Exo_hw_sw}a). The device had a total mass of $4.5$~kg and was capable of delivering joint torques up to $18$~Nm in the sagittal plane. Actuation was provided by two quasi-direct-drive brushless DC motors (AK80-9, CubeMars), one per hip joint. Mediolateral-axis angular velocity was measured by IMUs (ICM20948, TDK InvenSense, Japan) mounted on the distal thigh bars of the exoskeleton. The system was powered by a $22.2$~V, $3300$~mAh lithium-polymer battery (HRB). Onboard computation was handled by an embedded processor (Jetson Orin Nano, NVIDIA), on which the pre-trained student policy was converted to TensorRT format and executed at $100$~Hz. Two parallel processes were maintained on the co-processor to ensure stable sensor data acquisition and real-time inference without mutual interference (Fig.~\ref{Exo_hw_sw}b).

To account for differences in body mass between the generic simulation model ($74.5$~kg) and individual participants, the commanded torque amplitude was scaled by the ratio of the participant's body mass to the model mass:

\begin{equation}
    u_{\text{scaled}} = u_{\text{cmd}} \times \frac{m_{\text{subject}}}{74.5}
    \label{eq:torque_scaling}
\end{equation}

where $u_{\text{cmd}}$ is the raw torque command from the student policy and $m_{\text{subject}}$ is the participant's body mass in kilograms. In addition, the torque filter time constant was set to $\tau_{\mathrm{LPF}} = 0.15$~s on hardware, compared to $0.1$~s in simulation, to better attenuate measurement noise and inter-subject gait variability. Additionally, the rate of change for the control command was applied. The clipped command $u'_t$ is defined as $u'_t = \text{clip}(u_t, u_{t-1} - c, u_{t-1} + c)$, where $u_t$ is the raw command, $u_{t-1}$ is the previous command, and the rate limit constant is $c = 0.5$.

\begin{figure*}[t!]
    \centering
    \includegraphics[width=\textwidth]{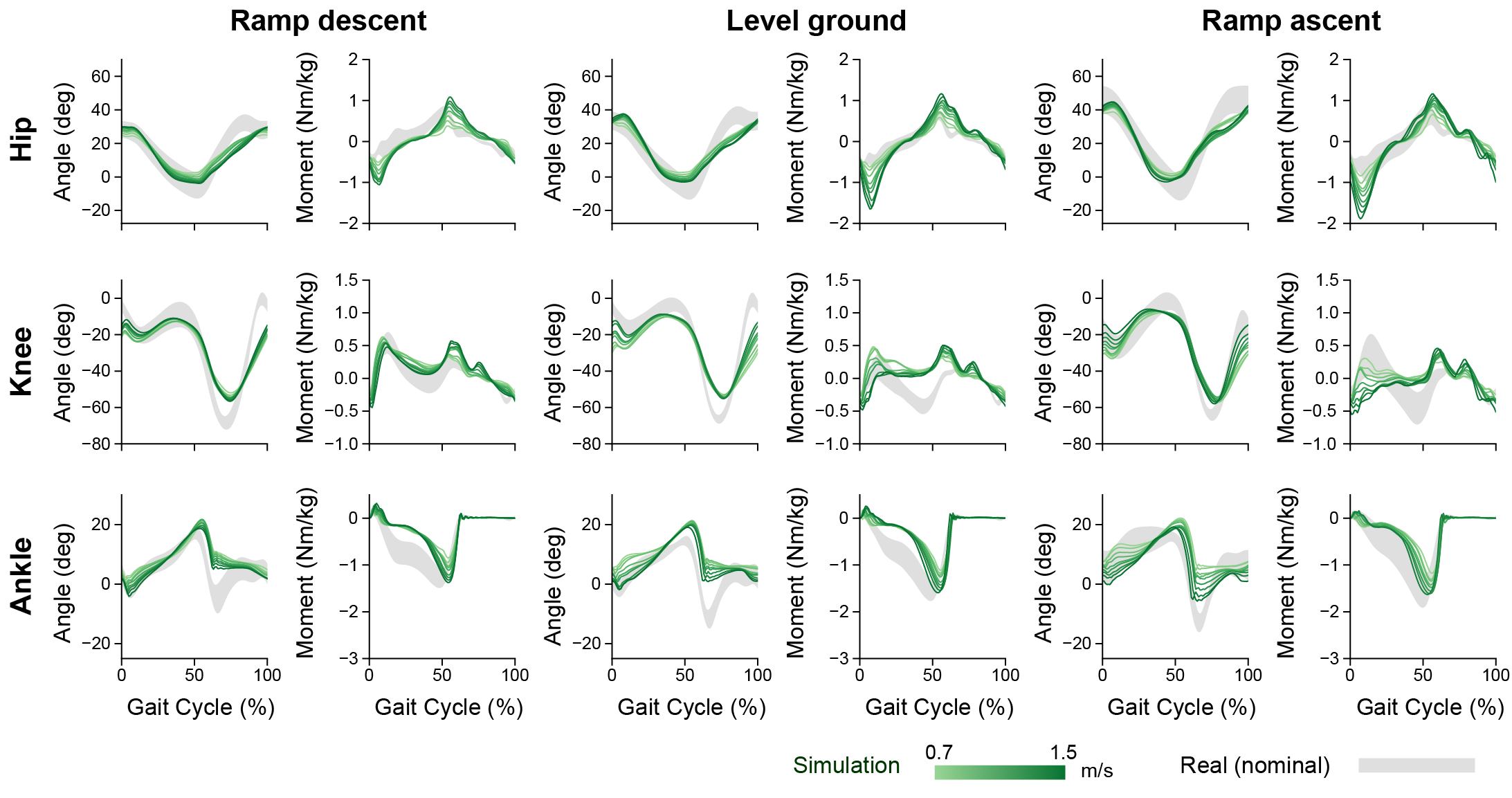}
    \caption{\textbf{Joint-level biomechanical comparison between simulation and human walking.}
    Sagittal-plane hip, knee, and ankle joint angles (left column of each slope) and net joint moments normalized by body mass (right column of each slope) are shown over the gait cycle for ramp descent ($-5^\circ$), level ground ($0^\circ$), and ramp ascent ($+5^\circ$).
    Colored curves denote simulation results across walking speeds (0.7--1.5~m/s), while the gray band indicates the mean $\pm$ standard deviation of open-source human experimental data~\cite{reznick2021lower}.
    All simulation traces correspond to the curriculum stage~2 no-exoskeleton condition, where exoskeleton torques were clamped to zero.}
    \label{sim_qual}
\end{figure*}
\begin{figure}[t!]
    \centering
    \includegraphics[width=\columnwidth]{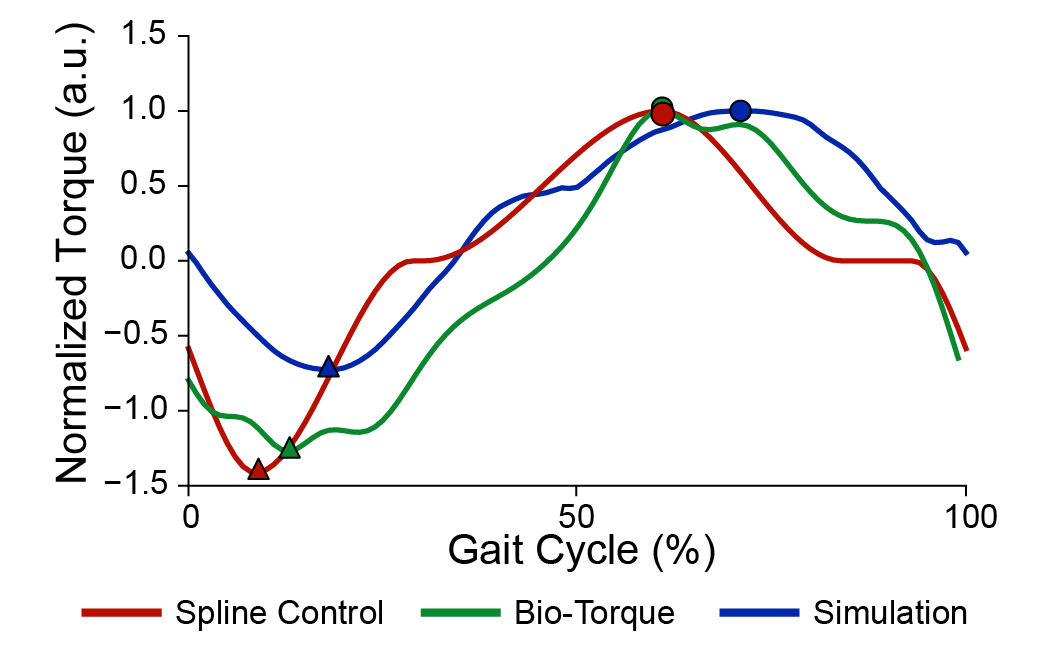}
    \caption{\textbf{Learned assistance torque profile compared with conventional controllers.} Representative hip assistance torque profiles over the gait cycle for a spline-based controller (Spline Control, red; $1.25$~m/s)~\cite{franks2021comparing}, a biological-torque-based controller (Bio torque, green; $1.2$~m/s)~\cite{molinaro2024task}, and the learned simulation policy (Simulation, blue; $1.2$~m/s). Each profile was normalized by its peak hip flexion torque magnitude for visualization and peak-timing comparison. Markers indicate the timing of the hip extension peak (minimum, triangles) and hip flexion peak (maximum, circles).}

    \label{sim_profile}
\end{figure}

\subsection{Human Experiment}

\subsubsection{Participants}
Five participants (4 male, 1 female; body mass $71.62 \pm 8.89$~kg, mean $\pm$ SD) were recruited for this study. All participants reported no history of neuromuscular or cardiovascular disorders. The study was conducted in accordance with the ethical principles of the Declaration of Helsinki and was approved by the Institutional Review Board of Carnegie Mellon University (STUDY2024\_00000506). Written informed consent was obtained from all participants prior to enrollment, including consent for the publication of identifiable information.

\subsubsection{Experimental Protocol}
Prior to the main experiment, participants donned the exoskeleton and completed a $5$-min acclimation walk on an instrumented split-belt treadmill (FIT5, Bertec, Columbus, OH, USA) at $1.2$~m/s with the exoskeleton powered on. The main trials were then conducted under three slope conditions: level-ground walking ($0^\circ$), ramp ascent ($+5^\circ$), and ramp descent ($-5^\circ$). Within each slope condition, treadmill speed was increased from $0.7$ to $1.5$~m/s in $0.2$~m/s increments, with each speed maintained for $30$~s, for a total duration of $3$~min $30$~s per condition. GRFs were recorded continuously from the treadmill for gait-event detection, and exoskeleton hip-torque commands were logged onboard throughout each trial. For each speed-slope condition, five consecutive gait cycles were extracted from the torque profiles and time-normalized to 101 points over the gait cycle.

\section{Results}
\subsection{Biomechanical Simulation Fidelity}

To quantify biomechanical consistency between simulation and human walking, we compared sagittal-plane hip, knee, and ankle joint angles and net joint moments from simulation against open-source experimental data~\cite{reznick2021lower} across three slope conditions (level ground, ramp ascent, and ramp descent), as shown in Fig.~\ref{sim_qual}. For quantitative comparison against the experimental dataset, we restricted the analysis to three representative speeds available in both datasets ($0.8$, $1.0$, and $1.2$~m/s), and all joint moments were normalized by body mass. Averaged across slopes and comparison speeds, angle RMSEs were $5.83 \pm 1.05^\circ$ (hip), $9.81 \pm 1.80^\circ$ (knee), and $6.32 \pm 0.92^\circ$ (ankle), and the corresponding moment RMSEs were $0.228 \pm 0.063$, $0.220 \pm 0.052$, and $0.344 \pm 0.044$~Nm/kg. Aggregated across all joints, slopes, and speeds, the global mean RMSE was $7.32 \pm 2.21^\circ$ for joint angles and $0.264 \pm 0.078$~Nm/kg for joint moments, with mean correlations of $r=0.845$ for angles and $r=0.761$ for moments.

\subsection{Learned Assistance Profile}
We compared the learned hip assistance torque profile against two established references: (1) a spline-based controller evaluated at $1.25$~m/s~\cite{franks2021comparing} and (2) a biological-joint-torque-based controller evaluated at $1.2$~m/s~\cite{molinaro2024task}, by quantifying the gait-cycle timing of the hip extension (minimum) and flexion (maximum) torque peaks (Fig.~\ref{sim_profile}). The simulation profile was obtained from level-ground walking at $1.2$~m/s. The biological torque control reference corresponds to the delayed profile reported by Molinaro \textit{et al.}~\cite{molinaro2024task}, constructed by averaging 20 gait cycles from each of 20 participants and applying a $125$~ms delay. The learned controller reached its extension peak at $17$\% of the gait cycle, compared with $9\%$ and $13\%$ for the spline and biological torque control profiles, respectively. The flexion peak occurred at $71$\% of the gait cycle for the learned profile, compared with $61$\% for both references.

\begin{figure}[t!]
    \centering
    \includegraphics[width=\columnwidth]{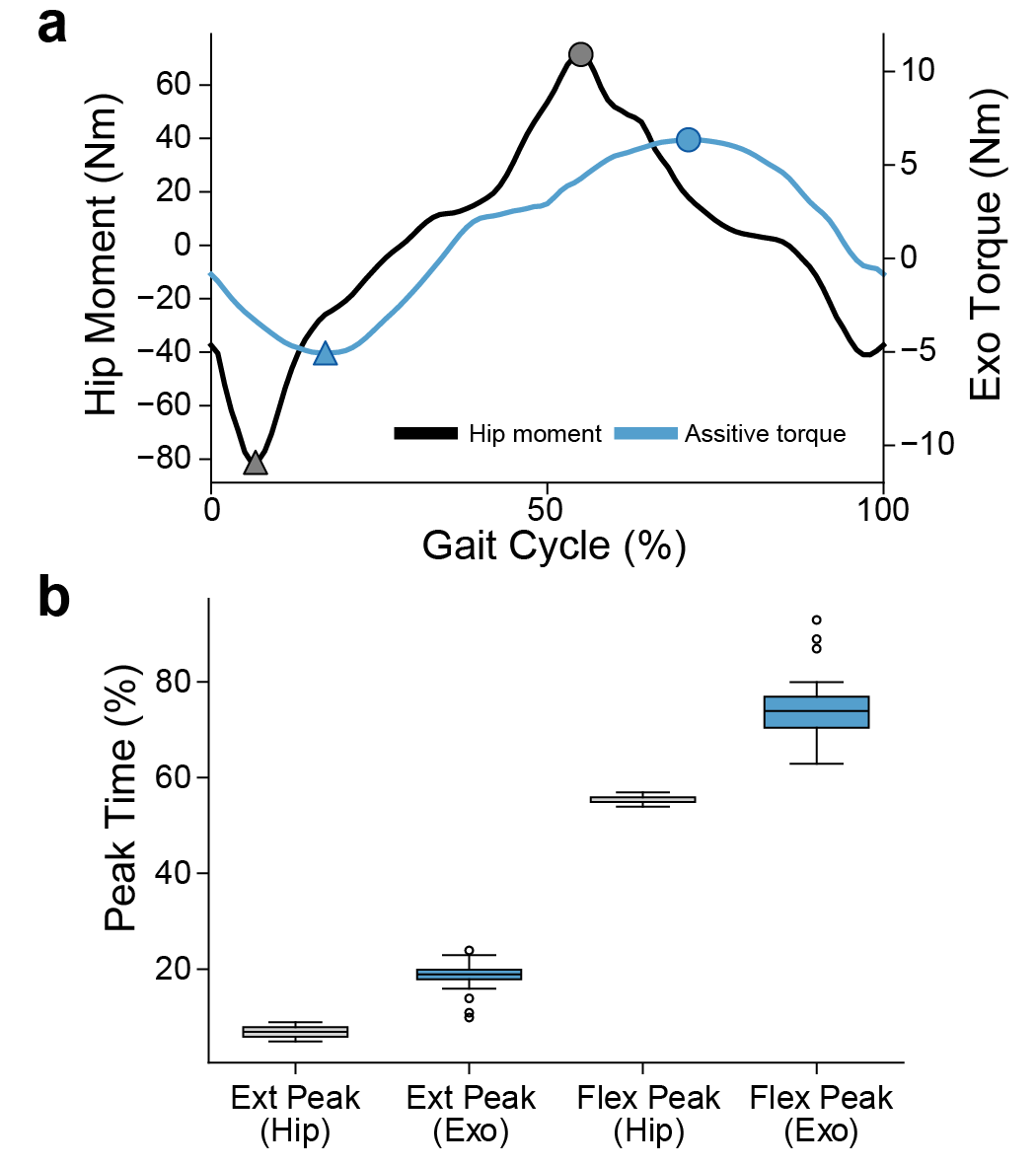}
    \caption{\textbf{Peak-timing comparison between the biological hip joint moment and the learned hip-exoskeleton assistance profile.} \textbf{(a)} Representative gait-cycle waveforms during level-ground walking ($0^\circ$, $1.2$~m/s), showing the sagittal-plane biological hip joint moment (black) and exoskeleton assistive torque (blue), with markers indicating the extension and flexion peaks. \textbf{(b)} Distribution of extension-peak and flexion-peak timings (percentage of the gait cycle) for the hip joint moment and exoskeleton torque across trials.}
    \label{delay_analysis}

\end{figure}

To further characterize the temporal relationship between biological joint demand and delivered assistance, we compared the gait-cycle timing of the sagittal-plane hip net joint moment peaks against the corresponding exoskeleton torque peaks (Fig.~\ref{delay_analysis}a). Averaged across conditions, the hip net joint moment reached its extension and flexion peaks at $6.9$\% and $55.4$\% of the gait cycle, respectively, whereas the exoskeleton torque peaked later at $18.6$\% (extension) and $74.1$\% (flexion). The resulting phase lag was $103$~ms for the extension peak and $166$~ms for the flexion peak.

\subsection{Effects of Exoskeleton in Simulation}
\begin{figure}[t!]
    \centering
    \includegraphics[width=\columnwidth]{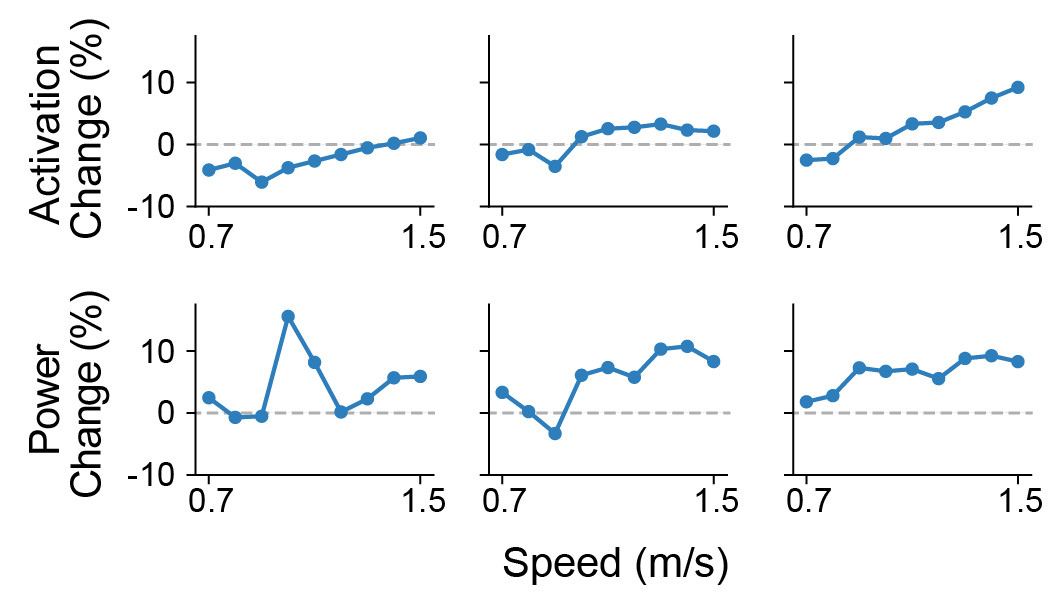}
        \caption{\textbf{Effect of exoskeleton assistance on muscle activation and positive joint power across conditions.} Summary metrics are shown as a function of walking speed for three slope conditions (ramp descent, level ground, and ramp ascent), related to no-exoskeleton conditions. The top row reports mean muscle activation averaged across all muscles over a $15$~s walking segment; the bottom row reports mean positive joint power averaged across bilateral hip, knee, and ankle joints in the sagittal plane.}

    \label{sim_effect}
\end{figure}

Fig.~\ref{sim_effect} summarizes the effect of hip exoskeleton assistance in simulation by comparing the no-exoskeleton and exoskeleton-assisted conditions across slopes and walking speeds. When averaged over speed, assistance reduced mean muscle activation by $1.1$\% on level ground and $3.4$\% on ramp ascent, and reduced mean positive hip-knee joint power by $6.3$\% and $7.0$\%, respectively. Ramp descent showed comparatively modest benefit, with mean muscle activation changing by $-2.2$\% and positive joint power reduced by $4.8$\%. To quantify how effectiveness scaled with walking speed, we pooled data across slopes and computed Pearson's correlation between walking speed and the percent reduction induced by assistance. The resulting correlations were $r=0.76$ for positive-power reduction and $r=0.98$ for mean-activation reduction, indicating that the benefit of exoskeleton assistance increased systematically with walking speed.

\subsection{Sim-to-Real Transfer of Assistance Profiles}
\begin{figure}[t!]
    \centering
    \includegraphics[width=\columnwidth]{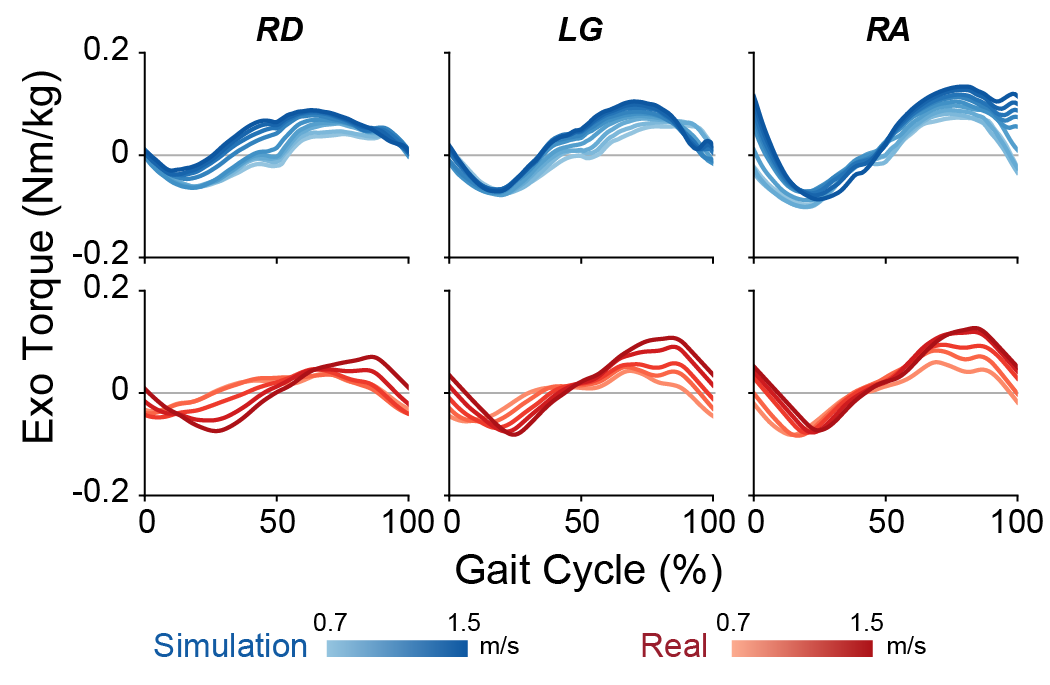}
    \caption{\textbf{Comparison of hip assistance torque profiles between simulation and hardware.} Gait-cycle-normalized right hip assistance torque is shown for ramp descent (RD), level ground (LG), and ramp ascent (RA). Blue curves denote simulated torque profiles across walking speeds from $0.7$ to $1.5$~m/s. Red curves show torque profiles generated by the distilled student controller on the physical device, evaluated at discrete speeds from $0.7$ to $1.5$~m/s in $0.2$~m/s increments. Color intensity increases with walking speed.}
    \label{sim_vs_real}
\end{figure}
To assess whether the assistance profiles learned in simulation were preserved upon hardware deployment, we compared right hip exoskeleton torque waveforms measured on the physical device against the corresponding simulation profiles over the full gait cycle. For each slope-speed condition, we computed Pearson's correlation coefficient ($r$) to quantify waveform-shape agreement and root-mean-square error (RMSE, Nm/kg) to quantify magnitude discrepancy. As shown in Fig.~\ref{sim_vs_real}, the assistance profiles exhibited good agreement across all conditions ($r=0.82\pm0.19$, RMSE $=0.03\pm0.01$~Nm/kg). Stratified by terrain, agreement improved progressively from ramp descent ($r=0.66\pm0.20$, RMSE $=0.04\pm0.01$~Nm/kg) to level ground ($r=0.83\pm0.16$, RMSE $=0.03\pm0.01$~Nm/kg) and ramp ascent ($r=0.98\pm0.01$, RMSE $=0.02\pm0.01$~Nm/kg), indicating that the learned assistance strategy was well preserved under the locomotor conditions best represented in the training distribution.

\section{Discussion}
This study establishes a sim-to-real pipeline for learning hip-exoskeleton assistance without motion-capture demonstrations or motion-mimicking objectives. Starting from a predictive neuromusculoskeletal walking simulation augmented with bilateral hip actuators, we trained a privileged reinforcement learning teacher policy across a range of walking speeds and slopes and subsequently distilled it into an IMU-only student policy for real-time onboard deployment. Using a two-stage curriculum, we directly compared the no-exoskeleton and exoskeleton assisted conditions, demonstrating that exoskeleton assistance reduced effort metrics in simulation while maintaining stable locomotion across task conditions. We benchmarked simulated sagittal-plane joint kinematics and kinetics against open-source human biomechanics data to confirm that the simulation remained biomechanically plausible across multiple slopes and speeds. We further compared the timing structure of the learned assistance profile against established reference profiles, revealing broadly consistent peak timing and waveform structure. Finally, we quantified sim-to-real transfer at the waveform level by comparing assistance torque profiles between simulation and hardware across matched speed-slope conditions. Together, these results support the feasibility of developing wearable-sensor exoskeleton controllers primarily in physics-based neuromusculoskeletal simulation, suggesting a pathway to substantially reduce experimental overhead during controller design when a biomechanically realistic simulation platform is available.

Because sim-to-real transfer is only meaningful if the simulator correctly predicts the direction of assistance effects, demonstrating benefits in simulation is a necessary prerequisite before expecting benefits on hardware. Terrain strongly modulates joint work demands: ramp ascent increases net positive work and elevates the hip's relative contribution to total positive power, whereas ramp descent is dominated by net negative work with limited hip involvement \cite{nuckols2020mechanics}. Consistent with these biomechanical demands, the learned policies reduced effort metrics on level ground and ramp ascent but provided minimal benefit on ramp descent, and the magnitude of reduction increased with walking speed. This speed dependence mirrors experimental observations that optimized assistance yields larger energetic benefits at faster walking speeds \cite{bryan2021optimized}, further supporting the use of simulation as a screening tool to identify conditions where real-world benefit is plausible prior to hardware evaluation.

Compared with experimentally driven approaches to exoskeleton controller development, including task-agnostic physiological state estimation, which requires large datasets with synchronized kinematics and GRFs \cite{molinaro2024estimating, molinaro2024task}, as well as human-in-the-loop optimization that relies on repeated, time-intensive treadmill experiments \cite{zhang2017human}, our approach front-loads controller development in physics-based neuromusculoskeletal simulation. Simulation enables rapid iteration, systematic comparison of design and training choices, and controlled evaluation across a wide range of walking speeds and slopes, reducing reliance on extensive motion-capture and force-plate data collection during the design phase while still yielding stable assistance across tasks. As a result, human experiments can be reserved primarily for validation and safety or efficacy testing, rather than serving as the primary foundation for controller discovery.

Although physics-based simulation has enabled strong sim-to-real transfer for autonomous legged robots, translating these successes to lower-limb assistive devices remains less common. In legged robotics, policies trained with large-scale simulated rollouts, curricula, and robustness techniques have been transferred to hardware with minimal additional tuning \cite{rudin2022learning, he2025asap}. For wearable exoskeletons, however, the controller must remain stable and comfortable while interacting with a human whose dynamics are redundant, subject-specific, and only partially observable, which raises the bar for both simulator fidelity and transfer validation.

A recent and notable example is the work by Luo \textit{et al.}, who learn a versatile hip-exoskeleton controller in simulation and report substantial metabolic reductions on hardware \cite{luo2024experiment}. Their framework incorporates a motion-imitation component trained on short motion-capture kinematic trajectories, which can stabilize multimodal locomotion learning but may also couple the learned strategy to the reference kinematics, potentially limiting flexibility when intentional kinematic deviations are desired, for example, in atypical or impaired gait. Conversely, our framework learns cyclic walking assistance without motion-mimicking objectives and complements hardware deployment with quantitative validation of joint-level biomechanical plausibility in simulation and waveform-level preservation of the learned assistance profile after sim-to-real transfer. Together with IMU-only policy distillation, these measurements provide an interpretable characterization of the remaining sim-to-real gap and clarify when physics-based neuromusculoskeletal simulation can serve as a reliable screening tool for exoskeleton controller design.

Exoskeleton control is commonly organized into three hierarchical tiers: high-level, mid-level, and low-level control \cite{kang2019effect}. The high-level module estimates locomotor state and user intent from onboard sensors and produces abstract assistance targets; the mid-level module converts these targets into implementable commands by selecting assistance timing, amplitude scaling, and waveform shape; and the low-level controller executes the resulting references on the actuators while maintaining electromechanical stability.

In practice, many mid-level controllers prescribe a phase-synchronized torque profile and tune its parameters using gait events and task context. These designs typically rely on manually specified peak locations or fixed temporal shifts to compensate for sensing and actuation delays; for example, Molinaro \textit{et al}. applied task-dependent constant delays on the order of $75$--$125$~ms to align assistance profiles across activities \cite{molinaro2024task}, and spline-based profile generators similarly encode predetermined peak timing within the gait cycle \cite{franks2021comparing}. Such approaches are interpretable and effective, but their timing logic is largely imposed \textit{a priori} rather than learned from interaction.

The distilled student policy, by contrast, maps a short history of thigh IMU gyroscope signals directly to hip torque commands, allowing assistance timing to emerge from the learned sensor-to-torque mapping. To interpret the resulting strategy, we compared the learned waveform against established reference profiles and performed a peak-timing analysis relative to the biological sagittal-plane hip net joint moment. Across conditions, the learned profile exhibited broadly consistent waveform structure and peak locations relative to prior paradigms \cite{molinaro2024task, franks2021comparing}, while its assistance peaks consistently lagged the corresponding biological hip-moment peaks by $103$~ms at extension and $166$~ms at flexion (Fig.~\ref{delay_analysis}). These delays are comparable in magnitude to previously hand-tuned shifts \cite{molinaro2024task}, yet they arise without prescribing a fixed offset, and the differing extension versus flexion lags indicate phase-dependent timing that would be difficult to capture with a fixed-delay heuristic.

This study was designed to establish and quantitatively validate a simulation-to-hardware learning pipeline; accordingly, the human evaluation was not intended as a rigorous efficacy study. The learned controller targets cyclic locomotion and has not been validated for non-cyclic or unstructured locomotion behaviors. The distilled policy relies on a single-axis thigh gyroscope history as its sole sensing modality, which enables practical onboard deployment but may be insufficient for broader locomotor contexts and could be sensitive to sensor placement variability. Beyond selecting the most consistent IMU axis for distillation, we did not explicitly optimize domain randomization and adaptation strategies to further reduce the sim-to-real gap. Future work should evaluate systematic approaches to narrowing this gap, including matching sensor noise and mounting dynamics in simulation, incorporating richer sensing modalities, and applying sim-to-real robustness objectives during distillation.

A promising direction for future development is extending this framework to impaired locomotion, where the clinical impact of exoskeletons may be greatest but exhaustive controller tuning through human experiments is often impractical. Collecting large, diverse datasets across tasks and environments is particularly challenging for patient populations, where repeated multi-condition trials are frequently constrained by fatigue, safety considerations, and logistical burden. Because our pipeline is grounded in predictive neuromusculoskeletal simulation rather than motion-capture demonstrations, it offers a pathway to incorporate neurophysiological constraints and internal control characteristics of pathological gait, rather than only matching observed kinematics. This capability could enable simulation-driven screening of assistance strategies and targeted evaluation of hypotheses about which interventions may improve stability, symmetry, or functional performance, followed by distillation to wearable sensing and targeted hardware validation. More broadly, incorporating richer task distributions, safety-aware training objectives, and subject-specific parameter adaptation within simulation may yield scalable sim-to-real controllers that generalize beyond steady-state locomotion and translate to clinically relevant deployment settings.

\section{Conclusion}
We presented a simulation-to-hardware pipeline for learning hip-exoskeleton assistance without motion-capture demonstrations or motion-mimicking objectives. We trained a privileged reinforcement learning teacher policy across a wide range of walking speeds and slopes within a predictive neuromusculoskeletal simulation and distilled it into an IMU-only student policy for real-time onboard deployment. Using a two-stage curriculum, we demonstrated measurable reductions in effort-related metrics. We validated simulation fidelity against open-source human biomechanics data and quantified sim-to-real transfer by comparing assistance torque waveforms across matched speed-slope conditions. Together, these results indicate that physics-based neuromusculoskeletal simulation can serve as a practical foundation for scalable exoskeleton controller development, reducing experimental burden during the design phase while providing interpretable, quantitative measures of sim-to-real preservation. Future work will extend this framework to broader locomotor contexts and clinical populations, with systematic evaluation of strategies to further narrow the sim-to-real gap.

\section*{Acknowledgment}
The authors would like to thank Dr. Jooeun Ahn, Eunsik Choi, and Jangwhan Ahn for contributions to the initial design concept of the muscle synergy model and generously sharing the experimental data that enabled the extraction of the muscle synergies. The authors also thank all subjects who participated in the study. 

\bibliographystyle{ieeetr}
\bibliography{references.bib}

@article{sawicki2020exoskeleton,
  title={The exoskeleton expansion: improving walking and running economy},
  author={Sawicki, Gregory S and Beck, Owen N and Kang, Inseung and Young, Aaron J},
  journal={Journal of neuroengineering and rehabilitation},
  volume={17},
  number={1},
  pages={25},
  year={2020},
  publisher={Springer}
}

@article{siviy2023opportunities,
  title={Opportunities and challenges in the development of exoskeletons for locomotor assistance},
  author={Siviy, Christopher and Baker, Lauren M and Quinlivan, Brendan T and Porciuncula, Franchino and Swaminathan, Krithika and Awad, Louis N and Walsh, Conor J},
  journal={Nature biomedical engineering},
  volume={7},
  number={4},
  pages={456--472},
  year={2023},
  publisher={Nature Publishing Group UK London}
}

@article{Gao2025,
  title = {Wearable technologies for assisted mobility in the real world},
  volume = {16},
  ISSN = {2041-1723},
  url = {http://dx.doi.org/10.1038/s41467-025-67126-4},
  DOI = {10.1038/s41467-025-67126-4},
  number = {1},
  journal = {Nature Communications},
  publisher = {Springer Science and Business Media LLC},
  author = {Gao,  Shuo and Chen,  Jianan and Xia,  Yunjia and Li,  Xuemeng and Ma,  Weihao and Yang,  Huixin and Li,  Jinchen and Zhou,  Xinkai and Jia,  Tianyu and Xu,  Yuchen and Uchitel,  Julie and Ta,  Dean and Qi,  Peng and Ge,  Junbo and Guo,  Yi and Qin,  Yajie and Kang,  Inseung and Xu,  Wenyao and Li,  He and Chang,  Jinke and Zuo,  Siming and Wang,  Shiwei and Luo,  Shan and Gionfrida,  Letizia and Hu,  Chen and Dong,  Shuqin and Guo,  Yongxin and Yuan,  Yixuan and Zhang,  Haixia and Chen,  Haotian and Pan,  Yu and Dai,  Chenyun and Ren,  Qinyuan and Loureiro,  Rui and Carlson,  Tom and Chen,  Wei and Zhang,  Yuanting and Kyriacou,  Panicos and Heidari,  Hadi and Nazarpour,  Kia and Prodromakis,  Themis and Casson,  Alexander and Makin,  Tamar R. and Cauwenberghs,  Gert and Farina,  Dario and Zhao,  Hubin},
  year = {2025},
  month = dec 
}

@InProceedings{Ma2025CoRL,
  title = 	 {Bipedal Balance Control with Whole-body Musculoskeletal Standing and Falling Simulations},
  author =       {Ma, Chengtian and Wei, Yunyue and Zuo, Chenhui and Zhang, Chen and Sui, Yanan},
  booktitle = 	 {Proceedings of The 9th Conference on Robot Learning},
  pages = 	 {4641--4656},
  year = 	 {2025},
  volume = 	 {305},
  url = 	 {https://proceedings.mlr.press/v305/ma25d.html},
}

@article{Chang2001,
  title = {Biomechanical simulation of manual lifting using spacetime optimization},
  volume = {34},
  ISSN = {0021-9290},
  url = {http://dx.doi.org/10.1016/S0021-9290(00)00222-0},
  DOI = {10.1016/s0021-9290(00)00222-0},
  number = {4},
  journal = {Journal of Biomechanics},
  publisher = {Elsevier BV},
  author = {Chang,  Chien-Chi and Brown,  Don R. and Bloswick,  Donald S. and Hsiang,  Simon M.},
  year = {2001},
  month = apr,
  pages = {527–532}
}

@article{Kang2025,
  title = {Online Adaptation Framework Enables Personalization of Exoskeleton Assistance During Locomotion in Patients Affected by Stroke},
  volume = {41},
  ISSN = {1941-0468},
  url = {http://dx.doi.org/10.1109/TRO.2025.3595701},
  DOI = {10.1109/tro.2025.3595701},
  journal = {IEEE Transactions on Robotics},
  publisher = {Institute of Electrical and Electronics Engineers (IEEE)},
  author = {Kang,  Inseung and Molinaro,  Dean D. and Park,  Dongho and Lee,  Dawit and Kunapuli,  Pratik and Herrin,  Kinsey R. and Young,  Aaron J.},
  year = {2025},
  pages = {4941–4959}
}

@article{Scherpereel2025,
  title = {Deep domain adaptation eliminates costly data required for task-agnostic wearable robotic control},
  volume = {10},
  ISSN = {2470-9476},
  url = {http://dx.doi.org/10.1126/scirobotics.ads8652},
  DOI = {10.1126/scirobotics.ads8652},
  number = {108},
  journal = {Science Robotics},
  publisher = {American Association for the Advancement of Science (AAAS)},
  author = {Scherpereel,  Keaton L. and Gombolay,  Matthew C. and Shepherd,  Max K. and Carrasquillo,  Carlos A. and Inan,  Omer T. and Young,  Aaron J.},
  year = {2025},
  month = nov 
}

@article{molinaro2024task,
  title={Task-agnostic exoskeleton control via biological joint moment estimation},
  author={Molinaro, Dean D and Scherpereel, Keaton L and Schonhaut, Ethan B and Evangelopoulos, Georgios and Shepherd, Max K and Young, Aaron J},
  journal={Nature},
  volume={635},
  number={8038},
  pages={337--344},
  year={2024},
  publisher={Nature Publishing Group UK London}
}

@article{ Geijtenbeek2019,
  author = {Thomas Geijtenbeek},
  title = {SCONE: Open Source Software for Predictive Simulation of Biological Motion},
  journal = {Journal of Open Source Software},
  year = {2019},
  volume = {4},
  number = {38},
  pages = {1421},
  publisher = {The Open Journal},
  doi = {10.21105/joss.01421},
  url = {https://doi.org/10.21105/joss.01421},
}

@misc{Geijtenbeek2021Hyfydy,
   author = {Geijtenbeek, Thomas},
   title = {The {Hyfydy} Simulation Software},
   year = {2021},
   month = {11},
   url = {https://hyfydy.com},
   note = {\url{https://hyfydy.com}}
}

@article{schumacher2025emergence,
  title={Emergence of natural and robust bipedal walking by learning from biologically plausible objectives},
  author={Schumacher, Pierre and Geijtenbeek, Thomas and Caggiano, Vittorio and Kumar, Vikash and Schmitt, Syn and Martius, Georg and Haeufle, Daniel FB},
  journal={iScience},
  volume={28},
  number={4},
  year={2025},
  publisher={Elsevier}
}

@inproceedings{chiu2025learning,
  title={Learning speed-adaptive walking agent using imitation learning with physics-informed simulation},
  author={Chiu, Yi-Hung and Lee, Ung Hee and Song, Changseob and Hu, Manaen and Kang, Inseung},
  booktitle={2025 International Conference On Rehabilitation Robotics (ICORR)},
  pages={835--841},
  year={2025},
  organization={IEEE}
}

@article{zhang2017human,
  title={Human-in-the-loop optimization of exoskeleton assistance during walking},
  author={Zhang, Juanjuan and Fiers, Pieter and Witte, Kirby A and Jackson, Rachel W and Poggensee, Katherine L and Atkeson, Christopher G and Collins, Steven H},
  journal={Science},
  volume={356},
  number={6344},
  pages={1280--1284},
  year={2017},
  publisher={American Association for the Advancement of Science}
}

@article{gunnell2025powered,
  title={Powered knee exoskeleton improves sit-to-stand transitions in stroke patients using electromyographic control},
  author={Gunnell, Andrew J and Sarkisian, Sergei V and Hayes, Heather A and Foreman, K Bo and Gabert, Lukas and Lenzi, Tommaso},
  journal={Communications Engineering},
  volume={4},
  number={1},
  pages={104},
  year={2025},
  publisher={Nature Publishing Group UK London}
}

@article{molinaro2024estimating,
  title={Estimating human joint moments unifies exoskeleton control, reducing user effort},
  author={Molinaro, Dean D and Kang, Inseung and Young, Aaron J},
  journal={Science robotics},
  volume={9},
  number={88},
  pages={eadi8852},
  year={2024},
  publisher={American Association for the Advancement of Science}
}

@article{kim2024soft,
  title={Soft robotic apparel to avert freezing of gait in Parkinson’s disease},
  author={Kim, Jinsoo and Porciuncula, Franchino and Yang, Hee Doo and Wendel, Nicholas and Baker, Teresa and Chin, Andrew and Ellis, Terry D and Walsh, Conor J},
  journal={Nature medicine},
  volume={30},
  number={1},
  pages={177--185},
  year={2024},
  publisher={Nature Publishing Group US New York}
}

@article{stable-baselines3,
  author  = {Antonin Raffin and Ashley Hill and Adam Gleave and Anssi Kanervisto and Maximilian Ernestus and Noah Dormann},
  title   = {Stable-Baselines3: Reliable Reinforcement Learning Implementations},
  journal = {Journal of Machine Learning Research},
  year    = {2021},
  volume  = {22},
  number  = {268},
  pages   = {1-8},
  url     = {http://jmlr.org/papers/v22/20-1364.html}
}

@article{reznick2021lower,
  title={Lower-limb kinematics and kinetics during continuously varying human locomotion},
  author={Reznick, Emma and Embry, Kyle R and Neuman, Ross and Bol{\'\i}var-Nieto, Edgar and Fey, Nicholas P and Gregg, Robert D},
  journal={Scientific Data},
  volume={8},
  number={1},
  pages={282},
  year={2021},
  publisher={Nature Publishing Group UK London}
}

@article{kang2025online,
  title={Online Adaptation Framework Enables Personalization of Exoskeleton Assistance During Locomotion in Patients Affected by Stroke},
  author={Kang, Inseung and Molinaro, Dean D and Park, Dongho and Lee, Dawit and Kunapuli, Pratik and Herrin, Kinsey R and Young, Aaron J},
  journal={IEEE Transactions on Robotics},
  year={2025},
  publisher={IEEE}
}

@article{luo2024experiment,
  title={Experiment-free exoskeleton assistance via learning in simulation},
  author={Luo, Shuzhen and Jiang, Menghan and Zhang, Sainan and Zhu, Junxi and Yu, Shuangyue and Dominguez Silva, Israel and Wang, Tian and Rouse, Elliott and Zhou, Bolei and Yuk, Hyunwoo and others},
  journal={Nature},
  volume={630},
  number={8016},
  pages={353--359},
  year={2024},
  publisher={Nature Publishing Group UK London}
}

@article{ha2025learning,
  title={Learning-based legged locomotion: State of the art and future perspectives},
  author={Ha, Sehoon and Lee, Joonho and van de Panne, Michiel and Xie, Zhaoming and Yu, Wenhao and Khadiv, Majid},
  journal={The International Journal of Robotics Research},
  volume={44},
  number={8},
  pages={1396--1427},
  year={2025},
  publisher={SAGE Publications Sage UK: London, England}
}

@article{franks2021comparing,
  title={Comparing optimized exoskeleton assistance of the hip, knee, and ankle in single and multi-joint configurations},
  author={Franks, Patrick W and Bryan, Gwendolyn M and Martin, Russell M and Reyes, Ricardo and Lakmazaheri, Ava C and Collins, Steven H},
  journal={Wearable Technologies},
  volume={2},
  pages={e16},
  year={2021},
  publisher={Cambridge University Press}
}

@inproceedings{rudin2022learning,
  title={Learning to walk in minutes using massively parallel deep reinforcement learning},
  author={Rudin, Nikita and Hoeller, David and Reist, Philipp and Hutter, Marco},
  booktitle={Conference on robot learning},
  pages={91--100},
  year={2022},
  organization={PMLR}
}

@article{he2025asap,
  title={Asap: Aligning simulation and real-world physics for learning agile humanoid whole-body skills},
  author={He, Tairan and Gao, Jiawei and Xiao, Wenli and Zhang, Yuanhang and Wang, Zi and Wang, Jiashun and Luo, Zhengyi and He, Guanqi and Sobanbab, Nikhil and Pan, Chaoyi and others},
  journal={arXiv preprint arXiv:2502.01143},
  year={2025}
}

@article{kang2019effect,
  title={The effect of hip assistance levels on human energetic cost using robotic hip exoskeletons},
  author={Kang, Inseung and Hsu, Hsiang and Young, Aaron},
  journal={IEEE Robotics and Automation Letters},
  volume={4},
  number={2},
  pages={430--437},
  year={2019},
  publisher={IEEE}
}

@article{kim2019reducing,
  title={Reducing the metabolic rate of walking and running with a versatile, portable exosuit},
  author={Kim, Jinsoo and Lee, Giuk and Heimgartner, Roman and Arumukhom Revi, Dheepak and Karavas, Nikos and Nathanson, Danielle and Galiana, Ignacio and Eckert-Erdheim, Asa and Murphy, Patrick and Perry, David and others},
  journal={Science},
  volume={365},
  number={6454},
  pages={668--672},
  year={2019},
  publisher={American Association for the Advancement of Science}
}

@article{osa2018algorithmic,
  title={An algorithmic perspective on imitation learning},
  author={Osa, Takayuki and Pajarinen, Joni and Neumann, Gerhard and Bagnell, J Andrew and Abbeel, Pieter and Peters, Jan},
  journal={Foundations and Trends{\textregistered} in Robotics},
  volume={7},
  number={1-2},
  pages={1--179},
  year={2018},
  publisher={Emerald Publishing Limited}
}

@article{dembia2020opensim,
  title={Opensim moco: Musculoskeletal optimal control},
  author={Dembia, Christopher L and Bianco, Nicholas A and Falisse, Antoine and Hicks, Jennifer L and Delp, Scott L},
  journal={PLOS Computational Biology},
  volume={16},
  number={12},
  pages={e1008493},
  year={2020},
  publisher={Public Library of Science San Francisco, CA USA}
}

@article{nuckols2020mechanics,
  title={Mechanics of walking and running up and downhill: A joint-level perspective to guide design of lower-limb exoskeletons},
  author={Nuckols, Richard W and Takahashi, Kota Z and Farris, Dominic J and Mizrachi, Sarai and Riemer, Raziel and Sawicki, Gregory S},
  journal={PloS one},
  volume={15},
  number={8},
  pages={e0231996},
  year={2020},
  publisher={Public Library of Science San Francisco, CA USA}
}

@article{bryan2021optimized,
  title={Optimized hip--knee--ankle exoskeleton assistance at a range of walking speeds},
  author={Bryan, Gwendolyn M and Franks, Patrick W and Song, Seungmoon and Voloshina, Alexandra S and Reyes, Ricardo and O’Donovan, Meghan P and Gregorczyk, Karen N and Collins, Steven H},
  journal={Journal of neuroengineering and rehabilitation},
  volume={18},
  number={1},
  pages={152},
  year={2021},
  publisher={Springer}
}

@article{slade2024human,
  title={On human-in-the-loop optimization of human--robot interaction},
  author={Slade, Patrick and Atkeson, Christopher and Donelan, J Maxwell and Houdijk, Han and Ingraham, Kimberly A and Kim, Myunghee and Kong, Kyoungchul and Poggensee, Katherine L and Riener, Robert and Steinert, Martin and others},
  journal={Nature},
  volume={633},
  number={8031},
  pages={779--788},
  year={2024},
  publisher={Nature Publishing Group UK London}
}

@article{ding2018human,
  title={Human-in-the-loop optimization of hip assistance with a soft exosuit during walking},
  author={Ding, Ye and Kim, Myunghee and Kuindersma, Scott and Walsh, Conor J},
  journal={Science robotics},
  volume={3},
  number={15},
  pages={eaar5438},
  year={2018},
  publisher={American Association for the Advancement of Science}
}

@article{witte2020improving,
  title={Improving the energy economy of human running with powered and unpowered ankle exoskeleton assistance},
  author={Witte, Kirby A and Fiers, Pieter and Sheets-Singer, Alison L and Collins, Steven H},
  journal={Science Robotics},
  volume={5},
  number={40},
  pages={eaay9108},
  year={2020},
  publisher={American Association for the Advancement of Science}
}

@article{kim2022reducing,
  title={Reducing the energy cost of walking with low assistance levels through optimized hip flexion assistance from a soft exosuit},
  author={Kim, Jinsoo and Quinlivan, Brendan T and Deprey, Lou-Ana and Arumukhom Revi, Dheepak and Eckert-Erdheim, Asa and Murphy, Patrick and Orzel, Dorothy and Walsh, Conor J},
  journal={Scientific reports},
  volume={12},
  number={1},
  pages={11004},
  year={2022},
  publisher={Nature Publishing Group UK London}
}

@article{lerner2017lower,
  title={A lower-extremity exoskeleton improves knee extension in children with crouch gait from cerebral palsy},
  author={Lerner, Zachary F and Damiano, Diane L and Bulea, Thomas C},
  journal={Science translational medicine},
  volume={9},
  number={404},
  pages={eaam9145},
  year={2017},
  publisher={American Association for the Advancement of Science}
}

@article{awad2017soft,
  title={A soft robotic exosuit improves walking in patients after stroke},
  author={Awad, Louis N and Bae, Jaehyun and O’donnell, Kathleen and De Rossi, Stefano MM and Hendron, Kathryn and Sloot, Lizeth H and Kudzia, Pawel and Allen, Stephen and Holt, Kenneth G and Ellis, Terry D and others},
  journal={Science translational medicine},
  volume={9},
  number={400},
  pages={eaai9084},
  year={2017},
  publisher={American Association for the Advancement of Science}
}

@article{pruyn2026portable,
  title={Portable hip exoskeleton improves walking economy for stroke survivors},
  author={Pruyn, Kai and Murray, Rosemarie and Gabert, Lukas and Foreman, K Bo and Lenzi, Tommaso},
  journal={Nature Communications},
  year={2026},
  publisher={Nature Publishing Group UK London}
}

@article{daniel1999learning,
  title={Learning the parts of objects by non-negative matrix factorization},
  author={Daniel, D},
  journal={Nature},
  volume={401},
  pages={788--791},
  year={1999}
}

@article{leem2026exo,
  title={Exo-Plore: Exploring Exoskeleton Control Space through Human-aligned Simulation},
  author={Leem, Geonho and Lee, Jaedong and Lee, Jehee and Song, Seungmoon and Won, Jungdam},
  journal={arXiv preprint arXiv:2601.22550},
  year={2026}
}

@article{badie2025bioinspired,
  title={Bioinspired morphology and task curricula for learning locomotion in bipedal muscle-actuated systems},
  author={Badie, Nadine and Al-Hafez, Firas and Schumacher, Pierre and Haeufle, Daniel FB and Peters, Jan and Schmitt, Syn},
  journal={Communications Engineering},
  volume={4},
  number={1},
  pages={115},
  year={2025},
  publisher={Nature Publishing Group UK London}
}

@article{bizzi2013neural,
  title={The neural origin of muscle synergies},
  author={Bizzi, Emilio and Cheung, Vincent CK},
  journal={Frontiers in computational neuroscience},
  volume={7},
  pages={51},
  year={2013},
  publisher={Frontiers Media SA}
}

@article{bernstein1967coordination,
  title={The coordination and regulation of movements},
  author={Bernstein, Nicholas},
  journal={(No Title)},
  year={1967}
}

\end{document}